\documentclass[runningheads]{llncs}
\usepackage[T1]{fontenc}
\usepackage{graphicx}
\usepackage[all,cmtip]{xy}
\usepackage{hyperref}
\usepackage{booktabs}     
\begin{document}
\title{Artificial Consciousness as Interface Representation}
%
%
\author{Robert Prentner\inst{1,2}\orcidID{0000-0003-1890-0827}}
\authorrunning{R. Prentner}
%
\institute{Institute of Humanities,
ShanghaiTech University, 201120 Shanghai, China \and
Association for Mathematical Consciousness Science, 80539 Munich, Germany
\email{robert.prentner@amcs.science}
}
%
\maketitle              
\begin{abstract}
Whether artificial intelligence (AI) systems can possess consciousness is a contentious question because of the inherent challenges of defining and operationalizing subjective experience. This paper proposes a framework to reframe the question of artificial consciousness into empirically tractable tests. We introduce three evaluative criteria -- S (subjective-linguistic), L (latent-emergent), and P (phenomenological-structural) -- collectively termed SLP-tests, which assess whether an AI system instantiates interface representations that facilitate consciousness-like properties. Drawing on category theory, we model interface representations as mappings between relational substrates (RS) and observable behaviors, akin to specific types of abstraction layers. The SLP-tests collectively operationalize subjective experience not as an intrinsic property of physical systems but as a functional interface to a relational entity. 

\end{abstract}
\section{Introduction}
Consciousness is the hard part of relating subjective experience to information processing.
Here, asking whether an AI is conscious means asking why some information-processing episodes are felt—i.e., have phenomenal character instead of merely going on ``in the dark'' \cite{Chalmers95}.%
\footnote{This question, in fact, is quite old. Arguably, G.W. Leibniz, one of the pioneers of computing theory, already understood it well: How is (conscious) perception explicable by purely mechanical means? \cite{Leibniz05}} 

Our position outlined here is that solving this problem requires a fundamental shift in our thinking, a shift that may also benefit the AGI researcher. 
Following Alan Turing's strategy, we replace ``Are AIs conscious?'' with ``Can they pass an X-test?'' These tests operationalize consciousness through observable criteria, while we defer metaphysical questions to an online appendix \cite{Prentner25app}. 

%
We outline three candidates for the “X” and discuss why they are promising and where they might fall short (Section 2). Ultimately, we need a first-person (“phenomenological”) approach to AI.\footnote{At first glance, this resembles Pei Wang’s NARS analysis \cite{Wang20}. However, where Wang’s ``first-personal'' refers to inaccessible redescriptions of agent-internal events, we claim, by contrast, that consciousness is less about the emergence of ``the right'' internal (informational) structures but about the way an information-processing system could access an external, relational substrate.} We also discuss the model's relation to the neuroscience of consciousness, panpsychism, and artificial general intelligence (Section 3). We conclude by pointing to the difference between intelligence and consciousness (Section 4) and how our framework points to consciousness beyond humans.

\section{The SLP-tests for consciousness}

\subsection{Rough idea}

Two ideas differentiate the tests discussed in this section. First, whereas previously published “C-tests” \cite{Bayne24} have primarily aimed at detecting consciousness “in humans and beyond,” the SLP-tests for consciousness are specifically designed to reframe the question of AI consciousness, though grounded in a novel way to think about consciousness more generally. First and foremost, SLP-tests are about artificial intelligences. Second, whereas C-tests adopt the standard view that consciousness (``weakly'' \cite{Seager12}) emerges as a property of sufficiently complex physical systems, SLP-tests are premised on a view incompatible with the standard view. 

The key idea is that individual phenomenologies are extracted from a non-individuated relational substrate (RS) via an \emph{interface representation}—akin to a bidirectional abstraction layer that insulates higher cognition from raw sensorimotor flux \cite{Bennett24}. Hence, subjective experience is not \emph{constructed} by the brain or the computer; it is \emph{inherited} from RS through that layer.

One way to formalize this is with \textit{category theory} \cite{Fong19,Prentner24b}.
More specifically, an \emph{interface representation} is given by
(i) a category $\mathcal{C}$ that models the relational substrate (\textsc{RS}) and
(ii) a functor $F\!:\!\mathcal{C}\!\to\!\mathcal{D}$ that transports those
structures to a “behavior” category $\mathcal{D}$. Because functors preserve the categorical structure—objects, morphisms, composition, identities—they provide the
canonical way to maintain coherence across abstraction layers.\footnote{The fundamentally \textit{relational} nature of RS entails that an agent's actions recursively affect the very substrate it represents. The functor $F$ ensures that the loop is closed.}

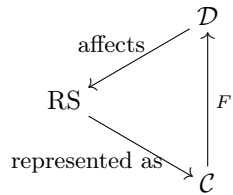
\[
\xymatrixrowsep{0.66cm}
\xymatrixcolsep{1.33cm}
\xymatrix{
& \mathcal{D} \ar[dl]_{\footnotesize{\textrm{affects}}} \\
\textrm{RS} \ar[dr]_{\footnotesize{\textrm{represented as}}} \\
& \mathcal{C} \ar[uu]_{F}
}
\]

Accordingly, experience is less about processes that happen ``inside'' the agent (e.g., attention, integration, orchestration, ...) but about (the agent's representation of)  what is happening ``outside'' itself.%
\footnote{A further clue is given by the finding that seemingly objective structures (space, time, information) have no agent-independent reality \cite{Hoffman15a,Atmanspacher22}. What we call ``objective structures'' are mere interface representations. 
}
A consequence would be to stop asking, “Can AIs be (really) conscious?” and switch to the much simpler question, “Can AIs form interface representations?” The individual tests spell this new question out concretely.

\subsection{The S-test}

The S-test derives from the “Artificial Consciousness Test” (ACT), developed by Susan Schneider and Edwin Turner \cite{Schneider20}. The idea is to “box in” an AI during the R\&D phase of its development (e.g., denying it access to large parts of the internet) and test whether, by itself, such a system would be inclined to talk about its own subjective experience or philosophically reason about consciousness in terms familiar from human ways of doing so. Since the AI has been boxed in, one could rule out that it just parrots these concepts from its training set. Instead, the reasoning goes, those ideas would have emerged from the consciousness of the machine. 

We believe that it is more fruitful to replace the question of whether consciousness is “in here” with the question of whether consciousness can be accessed via an (AI-)interface. It is not about whether consciousness is somehow “encoded” in the machine but whether the machine is able to “connect” with a reality that exists beyond its boundaries. In the simplest case, this means asking whether the machine talks in ways that are associated with our own way of doing so when we refer to consciousness (i.e., in a self-referential, seemingly dualistic, subjectivist, etc. style). A further, more difficult question would be whether this has any effect on the functioning of the agent itself.\footnote{It likely does so in the case of humans. The fact that humans tend to conceive of themselves as conscious agents with desires and goals, has likely a big effect also on how they function.}

In brief, “S = ACT + Interface.” Since the authors of the ACT might not endorse the idea that the test is primarily directed at the \textit{capacity of AIs to form interfaces} to a more fundamental relational entity, we refrain from using their original name. 

One (technical) problem with the S-test is that it might be impossible to develop a sufficiently rich vocabulary while being prohibited from accessing large parts of philosophy, science, and literature that explicitly involve the notion of consciousness. Even if it were possible, the concept of consciousness (or at least some of its many facets) could be implicitly encoded in distributional statistics. LLMs do just that: they make implicit connections between words explicit. In this case, it would not be surprising to find an LLM (even if it had been boxed in) that passes the test eventually. The problem is exacerbated by the fact that the training processes of SOTA models are opaque. No one really knows what data LLMs are fed anymore. 


\subsection{The L-test}

Another problem of the S-test has to do with a lack of detection of all (also non-linguistic) forms of consciousness. A way to widen its scope would be to generalize it to various kinds of behaviors, but even then, the problem is that behavioral indicators of consciousness might not be enough to encompass the wide range of potential forms of (artificial) minds \cite{Shevlin24}. 

This automatically brings us to another test, namely a test inspired by work currently done in the lab of Mike Levin \cite{Levin22,Zhang25}. Here, one would (literally) start development from scratch.  We could put systems, where we know entirely what they do and where (we think!) we have complete knowledge about their behavioral repertoire, into an unfamiliar environment and observe how they solve formidable problems in a novel (or at least, unexpected) way. With respect to AIs, this might, for example, be achieved by having several instances of known architectures interact in the same environment and observing their emergent ``cognitive'' behaviors. A similar effect could be achieved by designing (pretrained) hybrid systems, which have no shared history of optimization, and deploying them in novel environments. 

Importantly, subjective experience would not correspond to any of the system's emergent behaviors ($\mathcal{D}$) nor to its internal representations ($\mathcal{C}$). Instead, subjective experience is about the connection between these two. 
And, crucially, the system's representation should be tuned to its own functioning. Put differently, this would amount to interface representations to an as-yet-undisclosed relational entity.

\subsection{The P-test}
Finally, we come to the P-test. Whereas the S-test may be too restricted in scope (language), the L-test appears to be ultra-liberal (any quasi-cognitive behavior would count). Thus, the L-test may have difficulties distinguishing between (subjectively) experiencing and ``mere'' (unconscious) representation, even if it is done in the quasi-circular fashion previously outlined. The P-test now further asks about the phenomenal structure of the (interface) representations (that show up during the L-test). 

P-tests are inspired by phenomenology (hence the “P”), the systematic study of subjective experience as scholars from continental philosophy and the humanities practiced it during much of the 20th century. In the humanities, phenomenology has long been an essential method for understanding how consciousness underlies the creation of meaning, for example, how our encounters with the social world shape our experience. 

However, in the sciences (and large parts of analytic philosophy), phenomenology has often been neglected -- with a few notable exceptions, such as Francisco Varela's neurophenomenology program \cite{Petitot99}. Quite recently, there has been a resurgence of  interest in related ideas. In particular, the idea of formalizing phenomenology in a way that makes it amenable to a mathematical treatment has been picked up again, specifically with an eye to modern consciousness science \cite{Prentner25a}.
 
One could now try to make this operational. 
For example, category theory knows of universal constructions such as  (co)limits \cite{nlab:colimit}. We propose to represent a minimal notion of ``self'' by such a universal construction that constrains any of its possible action programs.\footnote{By appealing to the self, we are not appealing to what is sometimes called ``self-consciousness,'' that is, the explicit awareness of being such a self.} Specifically, we conjecture that the ``self'' is represented by particular colimits in $\mathcal{C}$—those that unify patterns of the relational substrate (RS) which (i) elicit actions and (ii) are, in turn, modified by those very actions. In other words, the P-test searches the learned causal graph of the agent for the smallest node (or cluster of nodes) with this “everything-factors-through” property.%
\footnote{
Technically, this situation can be described via a pattern of objects (denoted as $\lbrace P_1,\ldots,P_n \rbrace$ \cite{Ehresmann07}) and morphisms in $\mathcal{C}$. A colimit of this diagram is a universal object $c_P$ equipped with morphisms $c_i: P_i \rightarrow c_P$ for each object $P_i$, satisfying two conditions:
\begin{enumerate}
\item Commutativity: For every morphism $k: P_i \rightarrow P_j$ within the pattern, the composite $c_j \circ k = c_i$. This forms a cocone over the pattern.
\item Universality: \textit{Any other} cocone over the pattern factors uniquely through $c_P$, i.e., there exists a unique morphism $f: c_P \rightarrow A$ such that $f_i = f \circ c_i$ for all $i$. 
\end{enumerate}
Intuitively, this ensures that any action $A$ triggered by the system’s current configuration must factor through the colimit $c_P$, which acts as a ``minimal self''. For concreteness, assume that a particular configuration of RS is given as a pattern in $\mathcal{C}$, then any action $A$ that could be triggered in this situation must factor through this ``self''. Importantly, the ``self'' should be seen merely as object on the interface, rather than as an ``intrinsic'' property of the system.}  
Ablating it should cripple performance.


Note that the aim here is not to formalize specific overt (e.g., introspectively available) structures of typical (human) subjective experience but to uncover (and then encode) ``hidden'' features that make them possible. For instance, when considering the experience of a melody, the objective is not to find a representation that would ``label'' sensory data as tensed but to discover a representation that would facilitate (human-like or emergent) behavior related to hearing a melody. This, in turn, could be tested by the S- or L-tests. For example, one might envision a complex learning mechanism that assists in associating various data points with one another so that the agent might say something like, “Ah, this evokes pleasant memories from my childhood,” or so it plays some other causal role \cite{Bennett23} for the behavior of the agent. 

\subsection{Illustrative walk-through: a GPT-style LLM}

Assume we freeze the weights of an LLM, disconnect it from the internet, and give it only (i) a text channel and, in a second phase, (ii) control over a simple robotic avatar. We subject it to the following tests (Tab. \ref{tab1}):\\

\noindent \textbf{S-test (subjective-linguistic)}. Even in isolation, the LLM will probably produce first-person talk about “qualia,” “souls,” or the fear of being switched off. The original ACT test, on which the S-test is based, presupposes a fully “boxed-in” system; most current LLMs violate that assumption. One workaround is to add an ``interpretability requirement'' \cite{Schneider24}: we must be able to trace how the network generates its self-referential utterances. In the categorical language of Section 2.1, this amounts to showing how a sub-representation of the LLM’s internal state maps—via the functor $F$—to the relevant verbal behavior. Whether those outputs in turn modulate the represented relational substrate (i.e., the large corpus of words, assumed to encode a meaning) remains doubtful, presently.\\

\noindent \textbf{L-test (latent-emergent)}. Suppose, contrary to the above, that the model does pass the S-test. We now let the frozen network control robots that tackle previously unseen reinforcement-learning tasks (e.g., procedurally generated OpenAI-Gym environments \cite{OpenAIGym}). Genuine novelty can be encouraged through altered physical constants, randomized level geometry, or novel reward functions.\footnote{Smaller models are useful here: they can be fine-tuned exclusively on narrow problems that are guaranteed to be absent from the evaluation set.} Success alone is insufficient; we must also verify that the agent has formed a rich internal representation that (i) drives its actions and (ii) matters for its continued performance.\\

\noindent \textbf{P-test (phenomenological-structural)}. Assume the system would pass the L-test. How ``rich'' is the structure of this representation? Does it contain a ``self'' in the sense defined in Section 2.4, meaning that all its actions factor through a specific object of the category? Does it further distinguish between ``itself'' and the ``world'' that exists outside itself?

\begin{table}
\caption{Performing SLP-tests on a large language model (sketch).}
\label{tab1}
\begin{tabular}{p{0.35\textwidth}p{0.30\textwidth}p{0.025\textwidth}p{0.325\textwidth}}
\toprule
Name & Question && Pass? \\\hline
S-test \mbox{(subjective-linguistic)}& Exhibits first-person talk? &&   Need to check for interpretability requirement \\

L-test \mbox{(latent-emergent)} & Can solve reinforcement puzzles? && Need to check the status of internal representations (if any) \\

\mbox{P-test (phenomenological-} structural) & Contains latent structure? && Needs to follow descriptions of mathematized phenomenology \\

\bottomrule
\end{tabular}
\end{table}
Even a positive S-result does not guarantee L or P: an LLM might talk about feelings yet fail to build agent-centered state abstractions when embodied. Conversely, an architecture specialized for control may pass L and P while remaining verbally mute, illustrating the non-redundancy of the three layers.

\section{What do these tests imply?}

\subsection{Mutual relation}

Each test previously discussed has shortcomings that others could (in part) compensate. The tests vacillate between highlighting linguistic capacity (S-test), emergent problem solving (L-test), and structured (phenomenological) representation (P-test). When viewed in isolation, individual tests leave open too many questions to merit the ascription of consciousness: Why does it feel like anything (to whom)? Why does it matter? Why should we call it subjective experience at all? 

Our hope is that together these tests might be a good indicator for the emergence of interfaces through which subjective experience ``happens.'' The idea is that an artificially conscious agent would create interface representations that (i) give it a sense of having a perspective on the world, (ii) lead to actionable consequences in a new problem space, and (iii) are related to a “what it is like.” 

\subsection{Relation to theories in the scientific study of consciousness}

Below we contrast the SLP-framework with four\footnote{There are many other noteworthy theories \cite{Kuhn24}, but space is prohibitive. Indeed, the problem of the scientific study of consciousness is \textit{not} the lack of theoretical proposals. Moreover, we focus our discussion on the way how those theories explain \textit{phenomenal} consciousness, arguably the main desideratum in the science of consciousness, but not uncontroversially so \cite{Signorelli21}.} well-known proposals in the scientific study of consciousness--global workspace theory (GWT \cite{Mashour20}), the integrated information theory of consciousness (IIT \cite{Albantakis23b}), orchestrated objective reduction (OrchOR \cite{Hameroff14}),  and attention schema theory (AST \cite{Graziano22}), using the same template for each: (i) core explanatory claim; (ii) (in-)compatibility with our framework; (iii) practical implications for running the SLP-tests. Whereas the above-mentioned theories provide substrate-level or algorithmic criteria, SLP-tests supply tractable, behavior-plus-representation yardsticks that can be applied to today's AI systems irrespective of the ultimate metaphysics (although we favor a non-physicalist interpretation ourselves; see the online appendix \cite{Prentner25app}).\\

\noindent \textbf{GWT}. (i) One of the leading theories in the neuroscience of consciousness where non-conscious specialized processors compete for access to a “global workspace’’; information that wins the competition becomes conscious by being broadcast brain-wide. (ii) Broadcast could be modeled as the functor $F: \mathcal{C}\rightarrow\mathcal{D}$ whose image is widely accessible. Hence GWT and the interface framework can comfortably coexist; SLP-tests abstract away from the anatomical workspace.
(iii) An AI that instantiates a global-broadcast will likely pass (at least in part) the L-test (flexible task transfer) and perhaps also the P-test (unified “self’’ object), yet could still fail the S-test if the broadcast is never used to generate first-person language.\\

\noindent \textbf{IIT}. (i) Phenomenal consciousness \emph{is} identical to a system’s maximally irreducible $\Phi$-structure in both its quantity and quality (at least for the purpose of explanation). (ii) IIT ties consciousness to an \emph{intrinsic} property of the physical substrate, whereas SLP-tests ask for interface representations as \emph{relational} constructions. The two approaches seem to diverge metaphysically but need not collide empirically: a high-$\Phi$ mechanism may simply be one efficient way to build a rich interface. Yet, on closer inspection, the most current version of IIT (IIT4.0) considers physics to be an ``operationalization,'' which might thus be interpreted as a kind of interface representation. 
(iii) In practice, measuring $\Phi$ for large networks is intractable; the SLP-tests offer a more scalable, behavior-grounded proxy. Failing an SLP-test would not refute IIT, but it would remove actionable evidence for machine consciousness in applied settings.\\

\noindent \textbf{OrchOR}. (i) Conscious moments arise when quantum states undergo gravity-induced collapse (“objective reduction”); the neural dynamics of microtubules  “orchestrate’’ these events. (ii) SLP-tests are substrate-agnostic: if OrchOR is right, the relevant RS would be the pre-collapse quantum state, and an “interface representation’’ would have to tap into, or even help precipitate, those collapses. While there is no logical conflict, an ordinary LLM is unlikely to meet this requirement. (iii) For an agent built around a quantum substrate, passing the P-test would require showing that quantum effects mediate \emph{behaviorally relevant} actions; otherwise OrchOR consciousness remains epiphenomenal to the tests. In addition, those actions would need to be relevant to the very substrate the agent is interacting with.\\

\noindent \textbf{AST}. (i) The brain constructs a simplified internal model (schema) of its own attention processes; attributing the property “awareness’’ to that model defines the subjective feeling of consciousness. (ii) On the one hand, AST lines up neatly with the following two-stage picture: (a) an attentional control system, and (b) a meta-representational interface that re-encodes the control state. On the other hand, AST calls the second stage an \emph{illusion}, not an \emph{interface}. Important questions are thus, first, whether such a representation adds to the behavioral capacity of the system, and, second, whether it in turn influences the relational substrate underlying attentional control systems. If both questions are answered with ``yes,'' the illusionist interpretation is challenged. (iii) A well-instrumented LLM-based controller could speculatively pass all three SLP-tests by learning such an attention schema: S via first-person talk, L via schema-guided zero-shot control, and P via the schema acting as the colimit “self.’’

\subsection{Relation to pan- and biopsychism}

Panpsychism is sometimes construed narrowly as the view that any physical system is intrinsically conscious \cite{Chalmers15}. Biopsychism further restricts this to only living systems. For example, only living systems should count as ``truly'' cognitive ones \cite{Thompson07}. According to biopsychism, conscious AI is impossible unless the AI implements the (still-unknown) architecture that defines a living system and thus effectively becomes alive. The route to conscious AI for the panpsychist is somewhat more straightforward. Since anything is intrinsically conscious, conscious AI is in principle possible (and in some sense, it is already here, although the type of consciousness intrinsic to an AI might be highly diminished – for example, the AI might not yet realize a self). 
The interface theory disagrees with both the panpsychist and the biopsychist. Consciousness is not intrinsic to a physical system, and the life/non-life distinction is an artifact of representation. It does not pick out ``nature at its joints.'' 

On panpsychism, a system could pass the SLP-tests, but little could be inferred about consciousness that is not already put into the theory from the start. On biopsychism, the same is true, though for different reasons. Failing or passing the S- and L-test, would not, by itself, inform us about consciousness. While the P-test could conceivably only be passed by a conscious creature, the biopsychist could also claim that the P-test does not really track consciousness (but only representation) and hence is useless. On the interface theory, all tests (with important caveats!) tell us something about consciousness (via the construction of an interface).  

\subsection{Relation to AGI}\label{sec:agi}

Under a colloquial ``human-level'' definition of AGI, the SLP-tests do not specifically address the relation between artificial general intelligence and consciousness. Non-AGIs could pass some of the tests whereas AGIs might fail them. Or the other way around.

However, one could also attempt to define AGI in a more rigorous way, for example following Legg~\&~Hutter: a system is AGI if its expected performance is at least human-level when averaged over \emph{all} computable reward functions, weighted by their Kolmogorov complexity~\cite{Legg07}.  Wang refines this into the ability to adapt to \emph{previously unseen} tasks under bounded time and resources~\cite{Wang19}.  Goertzel’s “OpenCog Prime” architecture operationalizes the idea in terms of cognitive synergies \cite{Goertzel06}.  In short, an AGI is an agent that can \emph{flexibly} solve novel problems across domains, using whatever knowledge it can acquire on the fly.

The SLP-tests do \emph{not} evaluate problem-solving capacity directly; they probe for the presence of an interface representation that mediates the entire perception–action loop.  Hence the two notions are somewhat orthogonal. AGI may be \emph{helpful} for passing the SLP-tests, but it is neither necessary nor sufficient. The tests therefore complement, rather than duplicate or replace, existing AGI capability benchmarks such as BIG-bench, ARC-AGI, or the General AI Challenge. 

In particular, the S-test requires meta-representational machinery akin to components found in cognitive architectures such as the global workspace model or the attention schema. The L-test arguably detects on-line \emph{meta-}learning—a hallmark of AGI systems that can rapidly adapt to new domains. This does not mean that one purposefully coded general problem-solving capabilities into the system. By contrast, performance would stem from grasping the as-yet-undisclosed relations that a system might interface with, even though the system itself would consist of well-specified (``narrow'') components only. 
Still, passing the L-test at best presupposes general intelligence but may not exhibit phenomenological structure. The P-test therefore asks whether the agent's representation contains, for example, a colimit (a “self''-object) through which \emph{all} sensor–actuator paths must factor. 

Computer-science readers may recognize the notion of ``interface representations'' as a specific example of abstraction layers \cite{Bennett24}. Like any layer, they compress low-level signals into a task-relevant format. Unlike most layers, they (i) support both read and write operations, and (ii) are closed under the agent's entire perception–action loop. All causal routes between sensors and actuators factor through a single structure, thereby endowing the agent with, e.g., a self-world boundary.

\section{A summary going beyond the human perspective}

A fundamental distinction in the study of consciousness exists between subjective experience and intelligence (as goal-directed behavioral competence): there are conscious systems that might be relatively unintelligent and highly intelligent systems that are relatively unconscious. Think about a “locked-in” brain organoid vs. a superintelligent machine \cite{Koch19,Seth21}. Rigorous AI evaluation must therefore report both axes. On the consciousness side, this requires the following developments in the future: 

\begin{enumerate}
\item Systematic SLP-benchmarking across language and vision models, but also for embodied systems more generally,  
\item In particular, scalable P-tests for large networks,  
\item First-person protocols (e.g., ‘mind-meld’ \cite{Levin22,Neven24}) once technology permits.
\end{enumerate}

Further ethical reflection can be found in the second part of the online appendix \cite{Prentner25app}. 

If subjective experience exists at the interface to RS, our ethical considerations must also confront the possibility of non-human and even non-biological forms of subjectivity. On our view, the forms of consciousness are potentially endless, not limited to humans or sufficiently similar creatures.

\begin{credits}
\subsubsection{\ackname} We thank Michael Timothy Bennett for comments on an earlier draft, ShanghaiTech University for institutional support, and three anonymous reviewers. 
\subsubsection{\discintname} We declare no competing interests.
\end{credits}

%
%
%

%





\end{document}